%% file: paper.tex
\documentclass[]{phai}

\input{math_commands.tex}

\usepackage{booktabs}
\usepackage{url}
\usepackage{hyperref}
\usepackage{graphicx}
\usepackage{amsmath}
\usepackage{amssymb}
\usepackage{xcolor}
\usepackage{pifont}
\usepackage{gensymb}
\usepackage{colortbl}
\usepackage{multirow}
\usepackage{tabularx}
\usepackage{textcomp}
\usepackage{rotating}
\usepackage{adjustbox}
\usepackage{subcaption}
\usepackage{algpseudocode}
\usepackage{amsthm}

\newcommand{\labname}{Texas A\&M University}
\newcommand{\methodname}{MoRE}

\definecolor{oursrow}{rgb}{0.88,0.97,0.88}
\definecolor{baselinerow}{rgb}{0.98,0.89,0.89}
\definecolor{ablationrow}{rgb}{0.90,0.96,0.98}

\newlength{\taskthumbnailwidth}
\newlength{\taskthumbnailheight}
\newcommand{\tabrowlabel}[1]{%
  \begin{tabular}[c]{@{}l@{}}#1\end{tabular}%
}
\newcommand{\tabrowlabelmiddle}[1]{%
  \raisebox{\dimexpr 0.5\taskthumbnailheight - 0.5\height\relax}[0pt][0pt]{#1}%
}
\newcommand{\taskthumbnail}[1]{%
  \includegraphics[width=\taskthumbnailwidth, keepaspectratio]{#1}%
}

\title{Behavior Uncloning: Distilling Mode Redirection into Policy Weights without Inference-Time Steering}

\author[1,\dagger]{Hao Wang}
\author[1]{Jiuzhou Lei}
\author[1]{Dayou Li}
\author[2]{Bangya Liu}
\author[1]{Minghui Zheng}
\author[3]{Manling Li}
\author[3,4]{Ruohan Zhang}
\author[1]{Zhiwen Fan}

\affiliation[1]{\labname}
\affiliation[2]{University of Wisconsin}
\affiliation[3]{Northwestern University}
\affiliation[4]{Stanford University}

\contribution[\dagger]{Project Lead}

\abstract{
Behavior-cloned policies often learn multiple behavior modes from demonstration datasets, including modes that are unsafe or otherwise undesired at deployment.
For example, a policy trained on diverse handover demonstrations may learn to pass a knife blade-first.
Standard remedies such as data curation and inference-time steering either require access to the original demonstrations for full retraining or add substantial inference-time overhead.
To address this gap, we propose \textbf{\textit{MoRE}} (\underline{Mo}de \underline{Re}direction), which redirects policy rollouts toward desired behavior modes through a short ``uncloning'' step.
Specifically, MoRE distills the redirection signal from a temporary mode classifier into the policy weights to steer behavior. A retain loss balances this edit by preserving desired-mode competence, allowing the standalone policy to suppress unwanted modes with zero inference-time overhead.
Across eight simulated and real-world tasks, MoRE improves average deployment success rate (SR) by 44 percentage points over the original mixed-mode policy.
Among all compared adaptation and steering baselines, MoRE achieves the strongest SR and approaches the filtered-data retraining reference, while preserving task competence and inference speed.
MoRE also generalizes across robot policy backbones, including Diffusion Policy and the $\pi_{0.5}$ VLA, diverse task categories, and real-world deployments.
}

\date{\today}
\correspondence{Hao Wang: \email{haohw@tamu.edu}}

\metadata[Code]{\url{https://github.com/phai-lab/behavior-uncloning}}
\metadata[Project Page]{\url{https://behavior-uncloning.github.io/}}

\begin{document}

\maketitle

\begin{figure*}[!htb]
  \centering
  \includegraphics[width=1\textwidth]{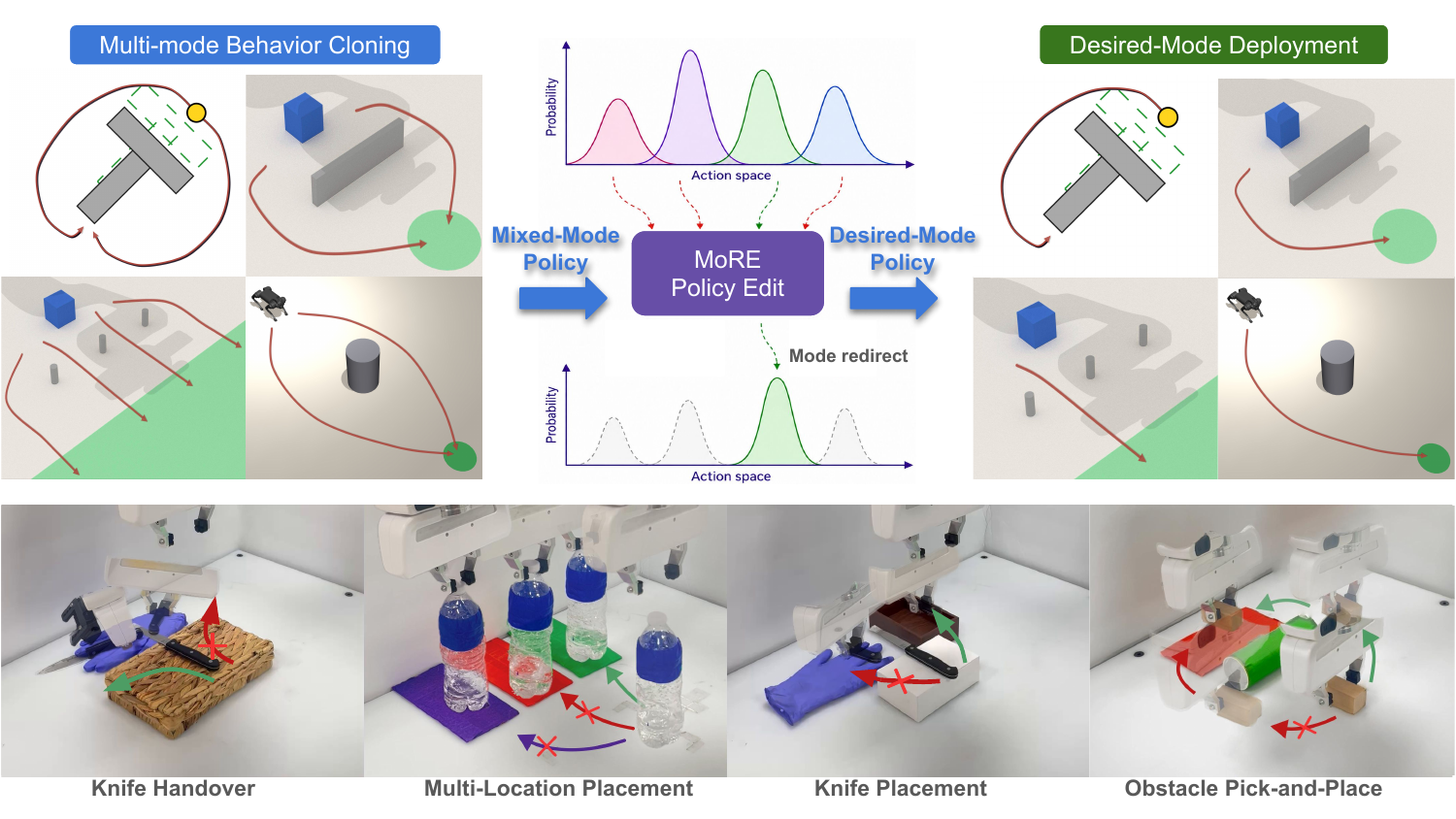}
    \caption{
    \textbf{\methodname{} edits a mixed-mode behavior-cloned policy into a standalone desired-mode policy without inference-time steering.}
    Top: Behavior cloning on multi-mode demonstrations yields a mixed-mode policy that can complete the task through multiple behavior modes, including modes that are undesired at deployment.
    MoRE performs a short policy-editing stage in which a mode classifier provides a training-time redirection signal for updating the policy weights toward desired behavior modes. After editing, the policy is deployed through the original inference path with no additional inference-time overhead.
    Bottom: four real-robot tasks from our evaluation suite.
    }
  \label{fig:teaser}
\end{figure*}

\section{Introduction}\label{sec:introduction}
Behavior cloning has driven significant advances in robot learning, where large and diverse demonstration datasets can produce capable visuomotor policies~\citep{droid, o2024open, walke2023bridgedata, mandlekar2018roboturk, gandhi2023eliciting}.
However, deploying these policies in real-world settings near people, such as homes, exposes a tension between dataset diversity and deployment requirements~\citep{HRIsurvey,gu2023human}.
Because large datasets aggregate demonstrations across different operators, scenes, and setups, they often contain multiple successful strategies, or behavior modes, for the same task~\citep{shafiullah2022behavior,jia2024towards}.
Different grasps or motion paths may all complete the task, yet some learned modes can be unsafe, undesirable to users, or incompatible with physical constraints at deployment.
For example, in the knife-handover task (Fig.~\ref{fig:teaser}), a policy trained on diverse demonstrations learns to pass the knife both handle-first and blade-first.
Although both modes can complete the handover, safe home deployment requires the handle-first mode.
Thus, deployment requires more than high task completion.
The policy must complete the task through an acceptable behavior mode, and the model should support explicit editing to suppress unsafe or undesired modes.

Filtered retraining provides direct data-centric mode control by removing undesired-mode demonstrations and training a new policy on the remaining data~\citep{mandlekar2021matters, belkhale2023data, hejna2024re}. However, it requires access to the original dataset and repeats costly training for each target mode.
Recent post-hoc methods instead leave the base policy unchanged and steer it during deployment. They re-rank sampled action candidates with external verifiers or value guidance~\citep{attarian2026update,pmlr-v270-nakamoto25a}, guide diffusion denoising with an auxiliary dynamics model~\citep{du2026dynaguide}, or intervene on VLA hidden states at each control step~\citep{haon2025mechanistic}. These methods can change behavior without full retraining, but the deployed system must run an additional steering module, verifier, or sampling loop, increasing inference cost and complicating low-latency robot control. These tradeoffs motivate behavior ``uncloning'', where mode-redirection signals are distilled into the policy weights so that undesired modes can be suppressed in a standalone edited policy without adding inference-time overhead.

To bridge this gap, we introduce \methodname{}, a post-hoc editing framework that distills a differentiable mode-redirection signal into the policy weights, steering behavior without adding inference-time cost. Given mode-labeled trajectories, MoRE trains a lightweight behavior-mode classifier and backpropagates a classifier-guided redirection loss through the policy. During editing, MoRE applies this signal only to undesired-mode samples in shared state-action regions. This focuses redirection where the policy can still change the subsequent behavior mode while avoiding disruption to later mode-specific execution. To balance mode suppression with task competence, a retain loss keeps desired-mode behavior close to the original policy. After editing, the classifier is discarded, and the edited policy runs through the original inference path while suppressing undesired modes.

Our primary contributions are:
\begin{itemize}
\item We formulate \textit{behavior uncloning}, an efficient policy-editing setting for suppressing deployment-undesired modes in already-trained robot policies, and evaluate it with deployment success rate, which counts a rollout as successful only when it both completes the task and follows an acceptable behavior mode.
\item We introduce \methodname{}, which distills classifier-guided mode redirection into
mixed-mode policy weights with a retain loss for task competence. It supports single- and multi-target edits, uses demonstrations or closed-loop rollouts, and removes the classifier after editing.
\item We validate \methodname{} across Diffusion Policy and the $\pi_{0.5}$ VLA backbone, binary and multi-way mode-control tasks, simulated and real-robot deployments, and manipulation and quadruped navigation settings. \methodname{} improves SR by an average of 44 percentage points over the original mixed-mode policy, outperforms compared adaptation and steering baselines, and approaches filtered-retraining references with no inference-time overhead.
\end{itemize}

\section{Related Work}
\label{sec:related_work}

\textbf{Behavior cloning and data-centric mode control.}
Behavior cloning (BC) has become a dominant recipe for robot policy learning, enabling capable visuomotor policies and increasingly generalist robot models from large demonstration datasets~\citep{florence2022implicit,chi2025diffusion,intelligence2025pi05visionlanguageactionmodelopenworld,shafiullah2022behavior,lee2024behavior,zhao2023learning, jia2024towards}. 
However, representing multiple modes is not the same as controlling which mode the policy executes at deployment. 
Some methods impose or discover mode structure during policy training~\citep{li2017infogail,eysenbach2018diversity,sharma2019dynamics,ajay2020opal, haarnoja2018latent, lynch2020learning,codevilla2018end}, but they address a different setting from post-hoc adaptation of a mixed-mode policy. 
Filtered retraining provides direct data-centric mode control: because BC is sensitive to demonstration composition and quality~\citep{mandlekar2021matters,belkhale2023data,kuhar2023learning}, dataset selection, weighting, and remixing have become common tools for shaping robot policy behavior~\citep{jang2022bc,brohan2022rt,o2024open,team2024octo,hejna2024re}. 
However, filtered retraining requires access to the original demonstrations and a full training run for each desired mode, while also discarding task-relevant information shared across modes. 
In contrast, \methodname{} starts from a mixed-mode policy and shifts its rollout behavior toward a specified target mode set.

\textbf{Inference-time steering and preference learning}.
Several post-hoc approaches avoid training a new filtered BC policy from scratch. 
Inference-time steering methods leave the policy unchanged and intervene during deployment, either by re-ranking sampled actions~\citep{attarian2026update,pmlr-v270-nakamoto25a}, guiding diffusion sampling~\citep{du2026dynaguide}, adding residual controllers~\citep{silver2018residual, johannink2019residual,abbas2023specialized,xiao2025self}, or steering hidden states in VLAs at each control step~\citep{haon2025mechanistic}. 
They avoid retraining the base policy, but the deployed system is no longer a single policy: it must be shipped with an external steering module. 
Preference-learning methods provide a natural post-hoc adaptation baseline
because they update policy weights from positive and negative trajectory sets~\citep{christiano2017deep,palan2019learning,pmlr-v119-brown20a,hejna2024contrastive,rafailov2023direct,ethayarajh2024kto}, with recent extensions to diffusion policies and robot policies~\citep{chen2025fdpp,xia2026human,hung2025nora, kim2024openvlaopensourcevisionlanguageactionmodel, intelligence2025pi05visionlanguageactionmodelopenworld, brohan2022rt, zitkovich2023rt}. 
In our setting, however, labels specify behavior modes rather than pairwise preferences alone: the goal is to move closed-loop rollouts into a target mode set while preserving task competence and the original inference interface. 
We include contrastive preference learning (CPL~\citep{hejna2024contrastive}) as a representative post-hoc preference baseline in Table~\ref{tab:task_suite}. 
More broadly, post-training weight edits in generative models and language models, such as concept erasure, guidance distillation, factual edits, and representation-level steering~\citep{gandikota2023erasing,gandikota2024concept,meng2022mass,heng2023selective,meng2023distillation,meng2022locating,zou2023representation,arditi2024refusal,dathathri2019plug, ho2022classifier} show that targeted behaviors can be removed or redirected directly in a model's weights. These limitations motivate a policy-editing view: instead of attaching steering computation at deployment, MoRE distills the mode-redirection signal into the policy weights, so deployment keeps the original single-policy inference path.

\section{Method}
\label{sec:Method}
\subsection{Problem formulation}
\label{sec:problem}
Behavior cloning trains a policy by supervised imitation of demonstrated observation-action pairs~\citep{pmlr-v9-ross10a,ross2011reduction}. For action-chunking policies, $a_t^i$ denotes the demonstrated action chunk associated with observation $o_t^i$. Let $\pi_\theta$ denote a policy parameterized by weights $\theta$, and let
\begin{equation}
    \mathcal{D}_{\mathrm{BC}}
    =
    \{\tau_i\}_{i=1}^{N},
    \qquad
    \tau_i=\{(o_t^i,a_t^i)\}_{t=1}^{H_i}
\end{equation}
denote the demonstrations used to train the original policy. The behavior-cloned policy is trained with the following optimization objective:
\begin{equation}
    \theta_0
    =
    \arg\min_{\theta}
    \frac{1}{\sum_i H_i}
    \sum_{i=1}^{N}\sum_{t=1}^{H_i}
    \mathcal{L}_{\mathrm{BC}}(o_t^i,a_t^i;\theta) .
\end{equation}
In our setting, each demonstration is annotated with a mode label, so the mixed-mode dataset can be partitioned as $\mathcal{D}_{\mathrm{BC}}=\bigsqcup_{k=1}^{K}\mathcal{D}_k$, where $\mathcal{D}_k$ contains demonstrations labeled as mode $k$.

\textbf{Behavior uncloning} starts from this mixed-mode policy and edits it for a desired deployment set \(S \subseteq \{1,\ldots,K\}\). For single-mode editing, \(S=\{m^\star\}\), where \(m^\star\) denotes the target behavior-mode index. For multi-target editing, \(S\) contains every acceptable mode.

The editing data $\mathcal{D}_{\mathrm{edit}}=\{x_j\}$ are represented at the
same granularity as the policy loss. Each sample $x_j=(o_j,a_j,m_j)$ contains an
observation, an action chunk, and a behavior-mode label. If
$\mathcal{D}_{\mathrm{edit}}$ comes from demonstrations, $a_j$ is the
demonstrated action; if it comes from closed-loop rollouts, $a_j$ is the action
executed by $\pi_{\theta_0}$. When only trajectory-level labels are available,
all samples from that trajectory inherit the same mode label.

Given the desired deployment set $S$, we partition the editing data into
\[
\mathcal{D}_{\mathrm{des}}
=
\{x\in\mathcal{D}_{\mathrm{edit}}:m(x)\in S\},
\qquad
\mathcal{D}_{\mathrm{undes}}
=
\{x\in\mathcal{D}_{\mathrm{edit}}:m(x)\notin S\}.
\]
Our goal is to find $\theta^\star$ such that $\pi_{\theta^\star}$ produces
closed-loop rollouts whose behavior modes fall in $S$, while preserving
task-level competence and requiring no additional inference-time overhead.

\subsection{Differentiable Mode-Redirection Editor}
\label{sec:mode_classifier}
Using the mode-labeled examples in $\mathcal{D}_{\mathrm{edit}}$, \methodname{}
first trains a $K$-way behavior-mode classifier on features produced by the
original mixed policy. For an example $x$, let $r_\theta(x)$ denote the policy-dependent classifier input. We train $g_\phi$ on cached features $r_{\theta_0}(x)$ to predict the mode label $m(x)$:
\begin{equation}
    z = g_\phi(r_{\theta_0}(x)) \in \mathbb{R}^K,
    \qquad
    q_\phi(m \mid r_{\theta_0}(x)) = \operatorname{softmax}(z)_m .
\end{equation}
After training, we freeze $\phi$. During editing, the same feature is recomputed
under the current policy as $r_\theta(x)$, so gradients from the frozen classifier
flow through $r_\theta(x)$ into the policy weights. The $K$-way output supports
both single-target and multi-target edits without retraining the classifier. The choice of $r_\theta(x)$ is policy dependent.
For VLA policies, we use $r_\theta(x)=h_{\theta,t}$, the mean-pooled last-layer
PaliGemma~\citep{beyer2024paligemma} prefix hidden state before the action expert.
For Diffusion Policy, we use $r_\theta(x)=(c_t,\hat a_{0,\theta})$, where $c_t$ is the policy-independent
observation condition and $\hat a_{0,\theta}$ is the predicted action
chunk reconstructed from the denoising output. In both cases, classifier training uses cached features from $\pi_{\theta_0}$,
while editing recomputes policy-dependent features under $\pi_\theta$ to provide
a differentiable redirection signal. For Diffusion Policy classifier features, diffusion steps are sampled uniformly from the policy-training noise schedule for both classifier training and editing. Additional implementation details are provided in the supplementary material.

\subsection{Optimization for Balancing Task Competence and Behavior Uncloning}
For a sample $x$, let $r_\theta(x)$ be the policy-dependent classifier input and
let $z=g_\phi(r_\theta(x))$ be the frozen-classifier logits. \methodname{} edits
the policy with a retain loss on desired-mode samples and a redirect loss on
gated undesired-mode samples:
\begin{equation}
    \mathcal{L}_{\mathrm{MoRE}}(\theta)
    =
    \underbrace{
    \mathbb{E}_{x \sim \mathcal{D}_{\mathrm{des}}}
    \left[
    \mathcal{L}_{\mathrm{BC}}(x;\theta)
    \right]
    }_{\text{retain}}
    +
    \gamma
    \underbrace{
    \mathbb{E}_{x \sim \mathcal{D}_{\mathrm{undes}}^{\mathcal{M}_{\theta}}}
    \left[
    \mathcal{L}_{\mathrm{red}}(x;\theta,\phi,S)
    \right]
    }_{\text{redirect}},
    \label{eq:method_loss}
\end{equation}
where
$\mathcal{D}_{\mathrm{undes}}^{\mathcal{M}_{\theta}}
=
\{x\in\mathcal{D}_{\mathrm{undes}}:x\in\mathcal{M}_{\theta}\}$
and $\mathcal{M}_{\theta}$ is the source-mode probability mask defined in
Sec.~\ref{sec:shared_state_masking}. The retain term uses the policy's original
imitation loss, e.g., noise prediction for Diffusion Policy and velocity
matching for flow-based VLAs.

For both single-target and multi-target editing, we use a unified
subset-probability redirection loss:
\begin{equation}
\mathcal{L}_{\mathrm{red}}(x;\theta,\phi,S)
=
-\log
\frac{\sum_{i\in S}\exp z_i}{\sum_{j=1}^{K}\exp z_j}.
    \label{eq:subset_redirection_loss}
\end{equation}
When $S=\{m^\star\}$, this reduces to the standard cross-entropy loss for the
target mode; when $|S|>1$, it redirects probability mass toward the acceptable
mode set.
Thus, the same loss redirects undesired samples either toward one target mode
or toward the acceptable mode set $S$. During editing, $\phi$ is fixed and gradients are taken only with respect to
$\theta$. We do not apply the imitation loss to undesired-mode samples, since it
would anchor the source mode while the redirect term pushes the policy toward
$S$. The classifier is discarded after editing, so deployment uses the original
inference path.

\subsection{Masking Redirection by Source-Mode Probability}
\label{sec:shared_state_masking}
Applying the redirect loss to every undesired-mode sample can be unstable.
We aim to edit undesired-mode samples only while they remain in regions shared
across behavior modes. After the rollout enters a mode-specific branch, the samples are strongly associated with their source mode. Redirection at this stage is less effective for changing the selected mode and more likely to degrade action
quality~\citep{laskey2017dart}.

We implement this mask using the classifier probability assigned to the sample's
source mode. For an undesired-mode sample $x_t$, let $m(x_t)\notin S$ denote its
source behavior-mode label. Given the current policy-dependent
classifier input $r_\theta(x_t)$, we apply redirection only to samples whose source-mode probability is below $\tau$:
\begin{equation}
    \mathcal{M}_{\theta}
    =
    \left\{
    x_t \in \mathcal{D}_{\mathrm{undes}}
    :
    q_\phi(m(x_t) \mid r_{\theta}(x_t)) < \tau
    \right\}.
    \label{eq:shared_state_mask}
\end{equation}
The mask is used only for sample selection and is evaluated using policy-produced features
$r_\theta(x_t)$, while the classifier parameters $\phi$ remain fixed. Thus, the redirect loss is applied only to undesired-mode samples whose source-mode probability is below $\tau$. In all experiments, we use $\tau = 0.5$, so an undesired-mode sample is redirected only when its
source mode receives less than half of the classifier probability mass.

Finally, we optimize Eq.~\ref{eq:method_loss} over the editable policy parameters: the trainable VLA parameters downstream of the frozen visual encoder,
including the PaliGemma transformer and action expert, and the denoising UNet for
Diffusion Policy. The classifier is discarded after editing, and the edited
checkpoint is deployed with the same inputs and inference path as
$\pi_{\theta_0}$.

\section{Experiments}
\label{sec:experiments}

We conduct experiments to answer four questions.
\textbf{Q1:} Can \methodname{} improve deployment success by redirecting closed-loop rollouts toward desired behavior modes while preserving task completion?
\textbf{Q2:} How closely does \methodname{} approach the filtered-data retraining reference, and how does it compare with preference learning and inference-time steering baselines?
\textbf{Q3:} Does \methodname{} require access to the original demonstrations, or can it edit a policy using only closed-loop rollouts from the mixed-mode policy?
\textbf{Q4:} Does \methodname{} generalize across policy backbones, mode counts, embodiments, and real-robot deployments?

\textbf{Experiment Settings.}
We evaluate \methodname{} on four simulated tasks in Table~\ref{tab:task_suite}: Push-T, two multi-mode tabletop manipulation tasks built in ManiSkill3~\citep{tao2024maniskill3}, and a quadruped navigation task built with MuJoCo Playground~\citep{2012mujoco, zakka2025mujoco}. We also evaluate on four real-robot tasks with a Franka Research 3 arm. Each task contains multiple behavior modes for the same task goal, examples are shown in Fig.~\ref{fig:teaser}. For Push-T, we reuse 96 demonstrations with clean mode patterns from the released Diffusion Policy trajectories~\citep{chi2025diffusion}. For the other three simulated tasks, we collect 60, 120, and 100 demonstrations, respectively. For the four tasks in real-robot experiments, we collect 40 demonstrations for each mode in a task. Push-T, Push-Wall, and Push-Pillars use state observations with Diffusion
Policy; Quadruped and real-robot Diffusion Policy tasks use image observations
encoded by a pretrained ResNet-18; and all $\pi_{0.5}$ experiments use images
encoded by the model's ViT encoder. During editing, we freeze all visual
encoders. For evaluation,
Diffusion Policy rollouts use 10-step DDIM sampling, while VLA rollouts use the
standard $\pi_{0.5}$ action-decoding procedure. We fix $\tau=0.5$ and choose
$\gamma\in\{0.002,0.005\}$ once per task/backbone setting, keeping it fixed
across target modes and evaluation seeds. MoRE edits use fewer than 500 gradient steps per target setting. Checkpoints are selected by held-out editing loss. For CPL~\citep{hejna2024contrastive} and DynaGuide~\citep{du2026dynaguide} baselines, we use the authors' released implementations and default hyperparameters.

\begin{table}[thb]
    \centering
    \scriptsize
    \setlength{\tabcolsep}{0pt}
    \renewcommand{\arraystretch}{1.10}
    \setlength{\arrayrulewidth}{0.25pt}
    \newlength{\tasklabelw}
    \newlength{\metriccolw}
    \setlength{\tasklabelw}{0.14\linewidth}
    \setlength{\metriccolw}{\dimexpr(\linewidth-\tasklabelw-1pt)/8\relax}
    \setlength{\taskthumbnailwidth}{\dimexpr 2\metriccolw\relax}
    
    \caption{
    \textbf{\methodname{} achieves the strongest SR across the Diffusion Policy benchmark with original-policy inference speed.}
    The visual row shows trajectories from closed-loop rollouts for each task.
    Within each SR or TCR column, entries are Avg/Max over target modes.
    For CPL, DynaGuide, and MoRE, each target-mode result is averaged over 3 training seeds and 50 initial states per seed (150 rollouts in total). The original mixed policy and filtered-retrain reference are evaluated with 50 rollouts per target mode.
    Original policy denotes the unedited mixed-mode policy. Push-T uses the mixed-mode DP checkpoint released by SFP~\citep{jiang2025streaming}, this is the same original-demonstration setting as Table~\ref{tab:pusht_assumptions}.
    Filtered-retrain ref. is not a post-hoc adaptation baseline. It denotes a strong
    retraining reference trained from scratch on mode-filtered demonstrations using
    ground-truth mode labels.
    }
    \begin{tabular}{
        @{}
        >{\raggedright\arraybackslash}m{\tasklabelw}
        !{\vrule width 0.25pt}
        >{\centering\arraybackslash}m{\metriccolw}
        >{\centering\arraybackslash}m{\metriccolw}
        !{\vrule width 0.25pt}
        >{\centering\arraybackslash}m{\metriccolw}
        >{\centering\arraybackslash}m{\metriccolw}
        !{\vrule width 0.25pt}
        >{\centering\arraybackslash}m{\metriccolw}
        >{\centering\arraybackslash}m{\metriccolw}
        !{\vrule width 0.25pt}
        >{\centering\arraybackslash}m{\metriccolw}
        >{\centering\arraybackslash}m{\metriccolw}
        @{}
    }
        \toprule
        Task
        & \multicolumn{2}{c}{Push-T}
        & \multicolumn{2}{c}{Push-Wall}
        & \multicolumn{2}{c}{Push-Pillars}
        & \multicolumn{2}{c}{Quadruped} \\
        \cmidrule(lr){2-3}
        \cmidrule(lr){4-5}
        \cmidrule(lr){6-7}
        \cmidrule(lr){8-9}
        
        \tabrowlabelmiddle{\tabrowlabel{Visual}}
        & \multicolumn{2}{c}{\taskthumbnail{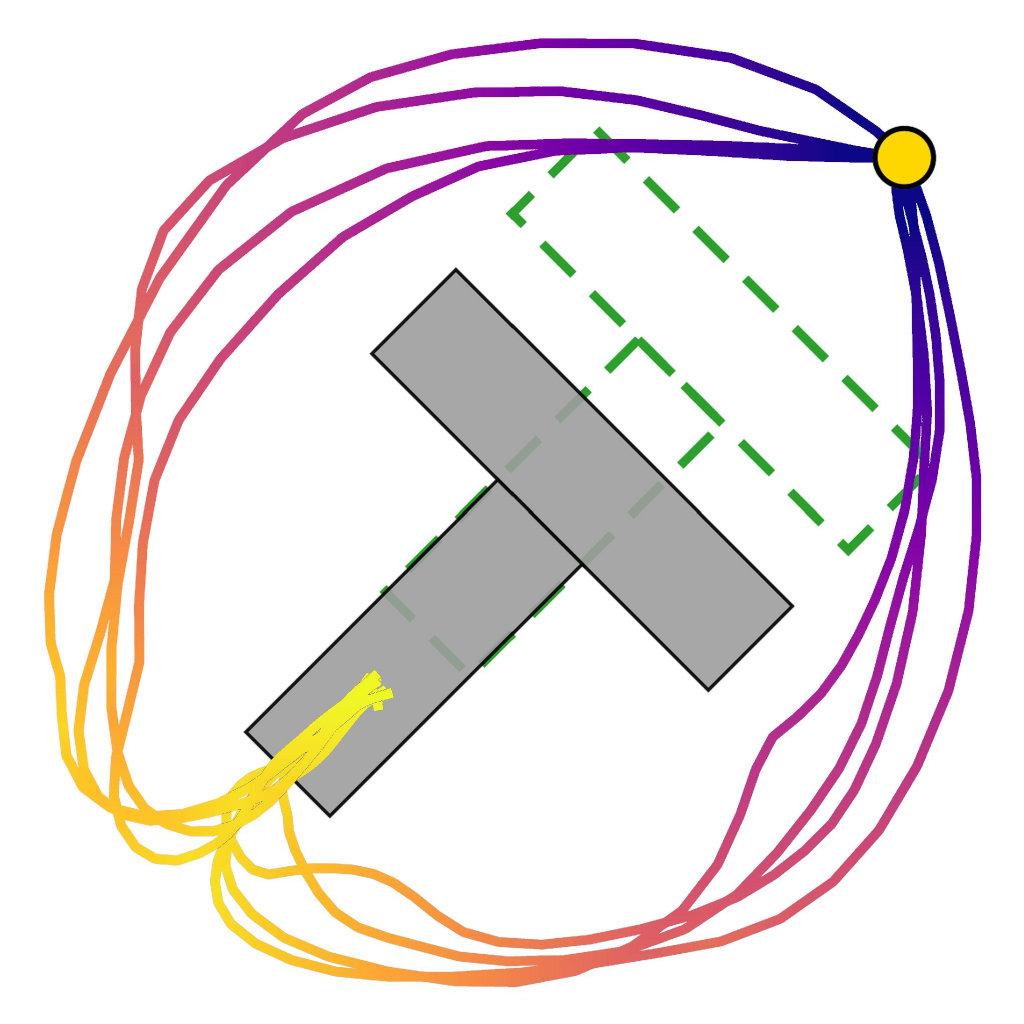}}
        & \multicolumn{2}{c}{\taskthumbnail{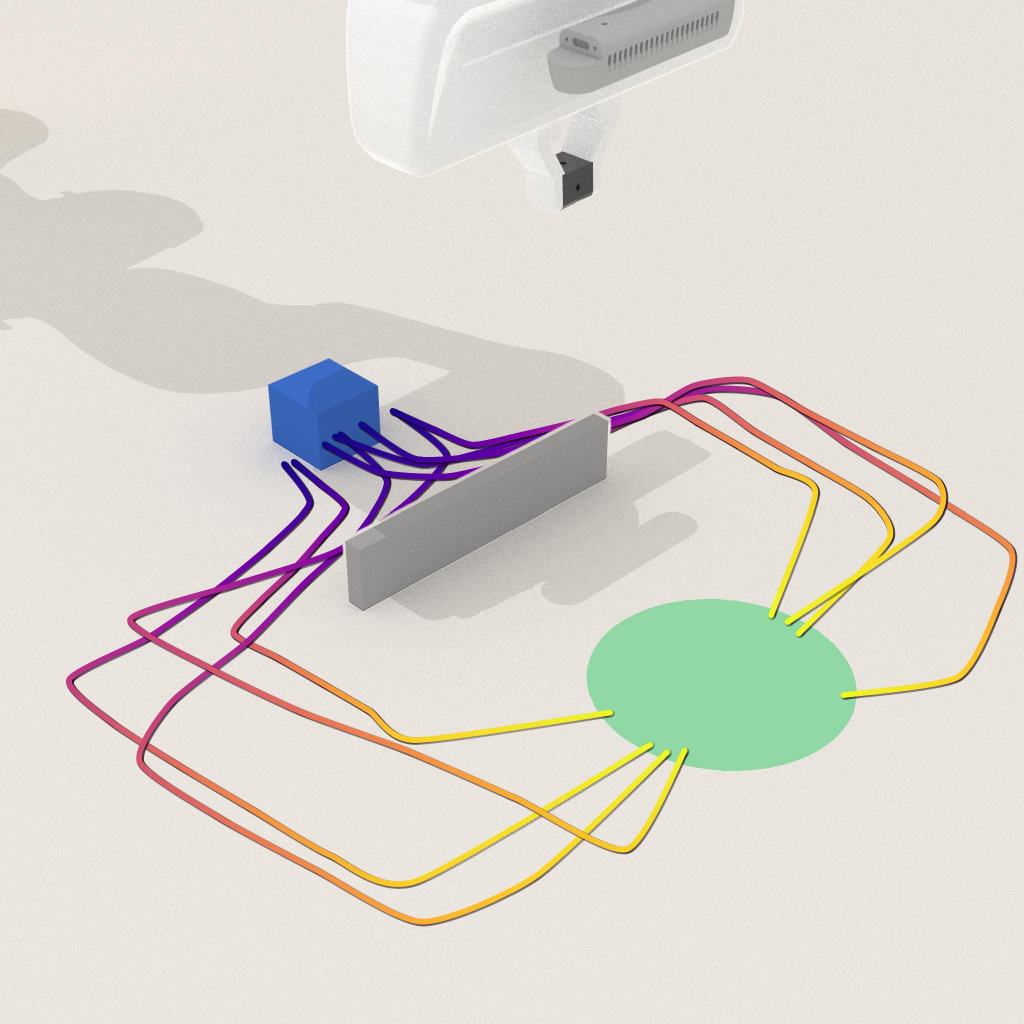}}
        & \multicolumn{2}{c}{\taskthumbnail{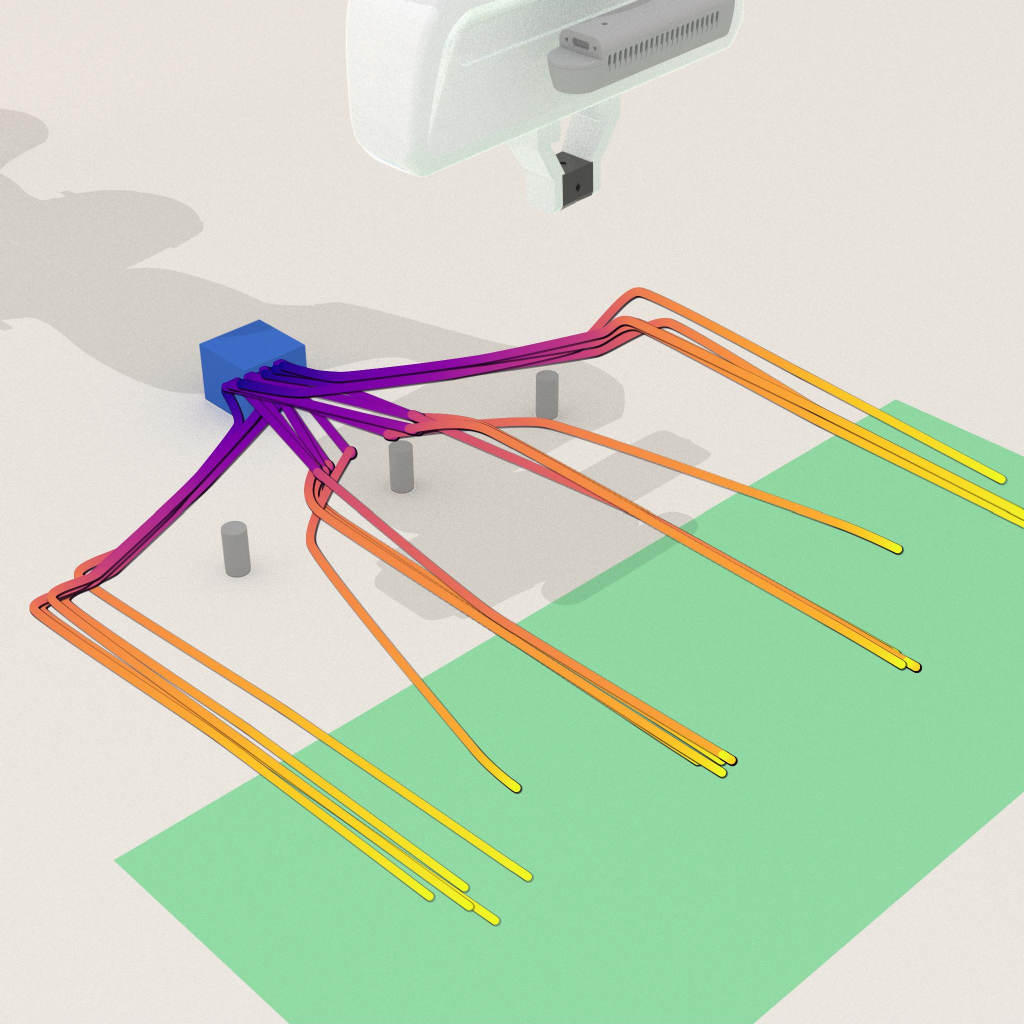}}
        & \multicolumn{2}{c}{\taskthumbnail{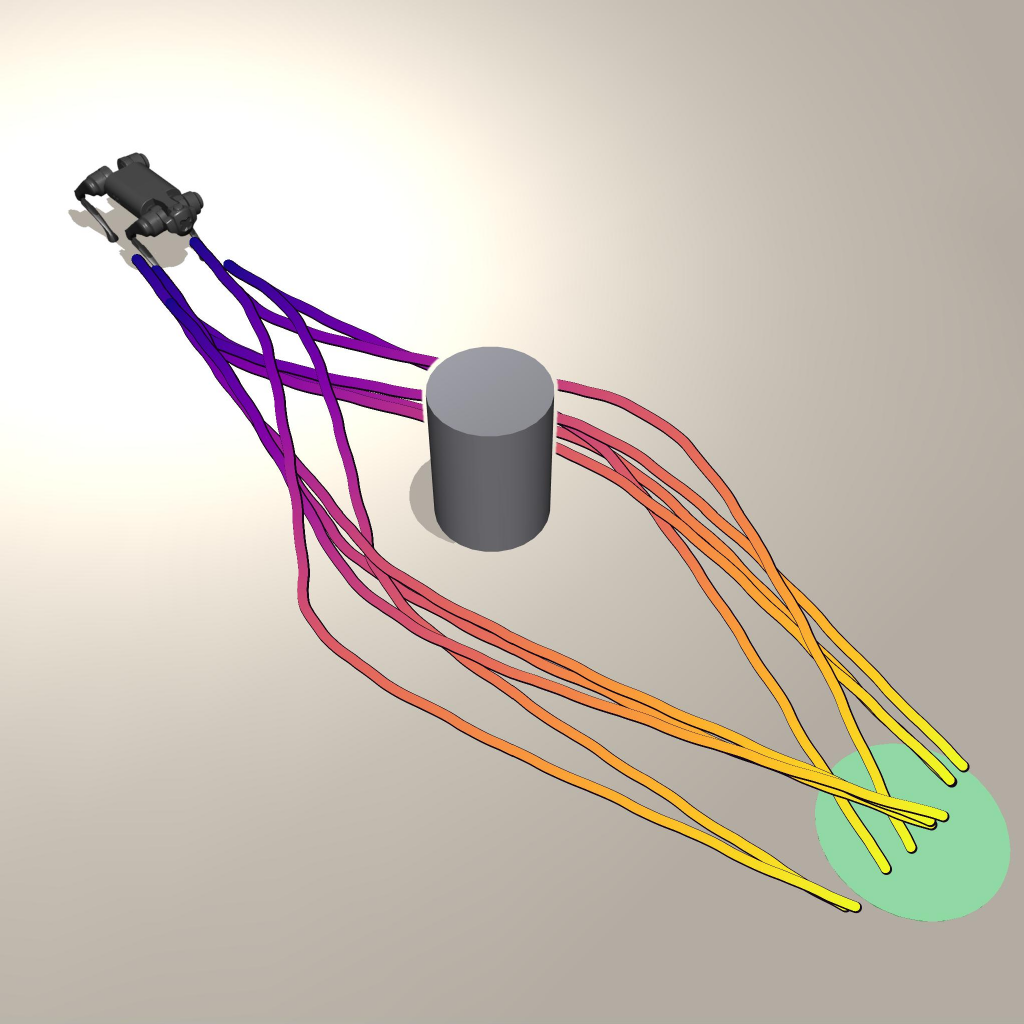}} \\

        \tabrowlabel{Modes}
        & \multicolumn{2}{c}{left or right}
        & \multicolumn{2}{c}{left or right}
        & \multicolumn{2}{c}{four routes}
        & \multicolumn{2}{c}{left or right} \\
        \midrule

        Metric(Avg/Max)
        & SR$ \uparrow$ (\%) & TCR$ \uparrow$ (\%)
        & SR$ \uparrow$ (\%) & TCR$ \uparrow$ (\%)
        & SR$ \uparrow$ (\%) & TCR$ \uparrow$ (\%)
        & SR$ \uparrow$ (\%) & TCR$ \uparrow$ (\%) \\
        \midrule

        \tabrowlabel{Original policy}
        & 50.0 / 54.0 & \textbf{100.0} / \textbf{100.0}
        & 37.0 / 38.0 & 74.0 / 74.0
        & 21.5 / 54.0 & \textbf{86.0} / 86.0
        & 50.0 / 56.0 & \textbf{100.0 / 100.0} \\

        \tabrowlabel{CPL}
        & 90.3 / 96.0 & 94.0 / 100.0
        & 50.7 / 58.7 & 53.0 / 61.3
        & 49.5 / 62.7 & 57.0 / 70.7
        & 67.7 / 77.3 & 80.7 / 98.0 \\

        \tabrowlabel{DynaGuide}
        & 49.7 / 52.7 & 100.0 / 100.0
        & 70.0 / 73.3 & 71.3 / 73.3
        & 34.3 / 45.3 & 61.5 / 66.0
        & 50.3 / 54.7 & 100.0 / 100.0 \\

        \rowcolor{oursrow}
        \tabrowlabel{\textbf{\methodname{}}}
        & \textbf{98.3 / 100.0} & 98.7 / \textbf{100.0}
        & \textbf{80.7 / 80.7} & \textbf{80.7 / 80.7}
        & \textbf{72.3 / 96.0} & 80.7 / \textbf{99.3}
        & \textbf{84.3 / 95.3} & 84.7 / 95.3 \\

        \midrule
        \tabrowlabel{Filtered-retrain ref.}
        & 100.0 / 100.0 & 100.0 / 100.0
        & 78.0 / 80.0 & 78.0 / 80.0
        & 72.5 / 98.0 & 75.0 / 98.0
        & 100.0 / 100.0 & 100.0 / 100.0 \\
        \midrule
        \multicolumn{9}{@{}l@{}}{
        \parbox{\linewidth}{
        \scriptsize
        \textit{Inference/adaptation cost.}
        DynaGuide performs inference-time guidance and runs at $8.47{\times}$ the
        original-policy latency. CPL and \methodname{} are post-hoc weight updates that
        deploy at the original-policy latency; \methodname{} uses fewer than 500 editing
        steps per target setting in our experiments. Filtered-retrain ref. trains a new
        policy from mode-filtered demonstrations and also deploys at the original-policy
        latency.
        }
        } \\
        \bottomrule
    \end{tabular}
    \label{tab:task_suite}
\end{table}
We evaluate deployment with success rate (SR) and task completion rate (TCR).
SR counts a rollout only if it both completes the task and follows a behavior
mode in the desired set $S$, whereas TCR ignores behavior mode. Failed
rollouts count toward neither metric, while completed non-target mode rollouts count
toward TCR but not SR. Rollout modes are assigned by task-specific geometry or
terminal regions in simulation, and by collection-time annotations on real robots. These labels are independent of the mode classifier.
Unless otherwise stated, simulated evaluations use 50 held-out initial-state
seeds per target mode, with rollout counts and repetitions specified in table
captions.

\textbf{\methodname{} improves deployment success rate by redirecting behavior modes. (Q1)}
Across the four Diffusion Policy tasks in Table~\ref{tab:task_suite}, \methodname{} improves average SR by 44 percentage points over the original mixed policy, including gains from 50.0\% to 98.3\% on Push-T and from 37.0\% to 80.7\% on Push-Wall. TCR remains high overall, indicating that MoRE mostly changes mode choice rather than simply trading off task competence.

\textbf{\methodname{} outperforms preference learning and steering baselines while approaching filtered-data retraining. (Q2)}
We compare against two levels of alternatives. Filtered-data retraining is a strong reference rather than a post-hoc baseline: it trains directly on mode-filtered demonstrations. It is not a strict theoretical upper bound, so \methodname{} can come close to this reference without retraining from scratch and even exceed it on some tasks. Among methods that start from the same mixed-mode policy, MoRE achieves the largest SR improvement. CPL~\citep{hejna2024contrastive} provides only partial mode shifts and can reduce TCR, while DynaGuide~\citep{du2026dynaguide} improves mode alignment on some tasks but requires inference-time guidance, causing an 8.47$\times$ inference-time slowdown on Push-Wall. In contrast, \methodname{} distills the redirection signal into the policy weights, so the final checkpoint deploys with the original sampler and no additional steering module. These results show that \methodname{} provides a stronger mode shift than generic preference learning or deployment-time guidance.

\textbf{Ablations.}
We ablate the loss components on Quadruped under the same editing setup as
\methodname{} and evaluate each ablated variant with 50 rollouts.
Full \methodname{} achieves 84.3\% SR and
84.7\% TCR, compared with 49.0\% SR / 100.0\% TCR for retain-only editing and
27.0\% SR / 53.0\% TCR for redirect-only editing. Thus, retaining desired-mode
behavior alone preserves task competence but does not redirect the mode, while
redirection alone degrades task completion. We also sweep the source-mode gate
threshold $\tau$ on Push-Wall with $\gamma=0.002$ over 3 seeds per target mode.
SR is stable for $\tau \in \{0.2,0.5,0.7\}$ at 80.3\%, 80.7\%, and 80.7\%,
whereas disabling the gate with $\tau=1.0$ reduces SR to 72.7\%. These ablations
support both the retain loss and gated redirection.

\begin{table*}[htb]
    \centering
    \footnotesize
    \renewcommand{\arraystretch}{1.08}
    \setlength{\arrayrulewidth}{0.25pt}
    
    \caption{
    \textbf{\methodname{} remains effective without access to the original demonstrations.}
    On Push-T, \methodname{} uses either selected original mixed-mode demonstrations (40 left, 56 right) or labeled closed-loop rollouts from the mixed policy (47 left, 53 right) as editing data.
    }
    \begin{tabular}{
        @{}l
        !{\vrule width 0.25pt}cc
        !{\vrule width 0.25pt}cc
        @{}
    }
        \toprule
        \multirow{3}{*}[-0.45ex]{Method}
        & \multicolumn{4}{c}{Push-T} \\
        \cmidrule(lr){2-5}

        & \multicolumn{2}{c}{\shortstack{Orig. Demos \\[-0.2ex] \scriptsize 40L + 56R}}
        & \multicolumn{2}{c}{\shortstack{Rollout \\[-0.2ex] \scriptsize 47L + 53R}} \\
        \cmidrule(lr){2-3}
        \cmidrule(lr){4-5}

        Metric (Avg / Max)
        & SR$\uparrow$ (\%) & TCR$\uparrow$ (\%)
        & SR$\uparrow$ (\%) & TCR$\uparrow$ (\%) \\
        \midrule

        Original policy
        & 50.0 / 54.0 & \textbf{100.0 / 100.0}
        & 48.0 / 52.0 & \textbf{96.0} / 96.0 \\

        \rowcolor{oursrow}
        \textbf{\methodname{}}
        & \textbf{98.3 / 100.0} & 98.7 / \textbf{100.0}
        & \textbf{88.0 / 98.0} & 88.0 / \textbf{98.0} \\

        \bottomrule
    \end{tabular}
    \label{tab:pusht_assumptions}
\end{table*}

\textbf{\methodname{} can use either demonstrations or mixed-policy rollouts as editing data. (Q3)}
In Table~\ref{tab:pusht_assumptions} we use the released mixed-policy checkpoint from Streaming Flow Policy~\citep{jiang2025streaming} and compare two sources of mode-labeled editing data.
The demonstration setting uses the clearest-mode trajectories from demonstrations in Diffusion Policy~\citep{chi2025diffusion}, while the rollout-only setting uses labeled closed-loop trajectories collected from the same mixed policy.
Demonstration-based editing reaches the strongest result, while rollout-only editing still improves SR to 88.0\%, showing that original demonstrations are useful but not strictly required.
When rollout modes can be reliably labeled, \methodname{} can edit a policy using trajectories generated by the mixed-mode policy itself.

\textbf{\methodname{} transfers across policy backbones, mode counts, embodiments, and real-robot deployment. (Q4)}
We evaluate transfer along four axes: (i) changing the policy backbone from
Diffusion Policy to a pretrained flow-matching VLA, (ii) increasing the number
of behavior modes and allowing multiple desired modes, (iii) transferring beyond tabletop manipulation to quadruped navigation, and (iv) moving from
simulation to real-robot deployments.

\begin{table}[hbt]
    \centering
    \scriptsize
    \setlength{\tabcolsep}{3pt}
    \renewcommand{\arraystretch}{1.10}
    \setlength{\arrayrulewidth}{0.25pt}
    \caption{
    \textbf{\methodname{} transfers to the $\pi_{0.5}$ VLA backbone and improves deployment success across both single-target and multi-target edits.}
    Mode Change denotes $K \rightarrow |S|$, where $K$ is the number of available behavior modes and $|S|$ is the size of the desired mode set.
    Within each SR or TCR column, entries are Avg / Max over target modes. 
    Each setting uses 50 evaluation seeds.
    For Push-Pillars, the $4 \rightarrow 2$ and $4 \rightarrow 3$ settings compute Avg / Max over the evaluated target subsets which have 2 and 3 desired modes, respectively. For multi-target edits, SR is computed by treating any rollout whose completed mode lies in the target set $S$ as mode-aligned.
    }
    \begin{tabular}{
        l
        !{\vrule width 0.25pt}cc
        !{\vrule width 0.25pt}cc
        !{\vrule width 0.25pt}cc
        !{\vrule width 0.25pt}cc
    }
        \toprule
        Task
        & \multicolumn{2}{c}{Push-Wall}
        & \multicolumn{6}{c}{Push-Pillars} \\
        \cmidrule(lr){2-3}
        \cmidrule(lr){4-9}
        Mode Change
        & \multicolumn{2}{c}{$2\rightarrow1$}
        & \multicolumn{2}{c}{$4\rightarrow1$}
        & \multicolumn{2}{c}{$4\rightarrow2$}
        & \multicolumn{2}{c}{$4\rightarrow3$} \\
        \cmidrule(lr){2-3}
        \cmidrule(lr){4-5}
        \cmidrule(lr){6-7}
        \cmidrule(lr){8-9}
        Metric (Avg/Max)
        & SR$\uparrow$ (\%) & TCR$\uparrow$ (\%)
        & SR$\uparrow$ (\%) & TCR$\uparrow$ (\%)
        & SR$\uparrow$ (\%) & TCR$\uparrow$ (\%)
        & SR$\uparrow$ (\%) & TCR$\uparrow$ (\%) \\
        \midrule
        \tabrowlabel{Original policy}
        & 41.0 / 52.0 & \textbf{82.0} / 82.0
        & 20.5 / 26.0 & 82.0 / 82.0
        & 41.0 / 48.0 & 82.0 / 82.0
        & 61.5 / 66.0 & 82.0 / 82.0 \\
        \rowcolor{oursrow}
        \tabrowlabel{\textbf{\methodname{}}}
        & \textbf{72.0 / 80.0} & 79.0 / \textbf{84.0}
        & \textbf{59.0 / 74.0} & \textbf{87.0 / 100.0}
        & \textbf{70.3 / 76.0} & \textbf{88.0 / 94.0}
        & \textbf{72.0 / 76.0} & \textbf{82.5 / 86.0} \\
        \midrule
        \tabrowlabel{Filtered-retrain ref.}
        & 82.0 / 88.0 & 82.0 / 88.0
        & 80.5 / 84.0 & 82.5 / 86.0
        & 72.3 / 88.0 & 74.7 / 88.0
        & 74.5 / 84.0 & 75.5 / 84.0 \\
        \bottomrule
    \end{tabular}
    \label{tab:vla_generalization}
\end{table}

\textbf{Transfer to VLA backbone. (Q4.i)}
We also apply \methodname{} to $\pi_{0.5}$, a pretrained vision-language-action model finetuned on our task data before editing. This setting differs from the compact Diffusion Policy benchmark in both representation and optimization: the mode classifier is attached to the mean-pooled last-layer PaliGemma prefix hidden state, and the edit updates the trainable VLA parameters used during task finetuning. The visual encoder is kept frozen during editing. This setup also introduces possible mode biases inherited from pretraining, making it a qualitatively different test of the same classifier-guided editing
objective. Nevertheless, \methodname{} improves SR across VLA tasks. On Push-Wall, \methodname{} increases SR from 41.0\% to 72.0\%. On Push-Pillars, it improves SR across both single- and multi-mode target settings. We further evaluate $\pi_{0.5}$ on real-world knife handover, where \methodname{} shifts the policy toward the desired handover mode while keeping TCR comparable to the finetuned mixed policy as shown in Table~\ref{tab:real-robot_dp}. These results suggest that \methodname{} is not tied to the Diffusion Policy denoising interface.

\textbf{More modes and multi-target edits. (Q4.ii)}
\methodname{} is not restricted to binary mode suppression. On four-route Push-Pillars in Table~\ref{tab:task_suite}, \methodname{} increases Diffusion Policy SR from 21.5\% to 72.3\%, showing that the edit can select a desired mode from more than two possible
behavior modes. We further evaluate multi-target mode editing with the VLA backbone, where the desired set can contain multiple acceptable modes rather than a single target mode. As shown in Table~\ref{tab:vla_generalization}, on Push-Pillars with $\pi_{0.5}$, \methodname{} improves SR from 41.0\% to 70.3\% in the $4 \rightarrow 2$ setting and from 61.5\% to 72.0\% in the $4 \rightarrow 3$ setting. These results show that the multi-target redirection objective can suppress undesired modes while retaining multiple desired modes.
The real-world bottle-placement task in Table~\ref{tab:real-robot_dp} exhibits the same capability: \methodname{} successfully preserves either the \{left, right\} or the \{middle, right\} subset of placement modes when there are three total modes.

\textbf{Transfer beyond tabletop manipulation. (Q4.iii)}
For the quadruped experiment, we first train a low-level Unitree Go1 joystick gait controller in the MuJoCo Playground environment~\citep{zakka2025mujoco} using Brax PPO~\citep{freeman2021brax,schulman2017proximal}. We freeze it and then train and edit a Diffusion Policy that outputs high-level body-frame velocity/yaw action chunks, where each action is $(v_x, v_y, \omega)$. The frozen gait controller tracks these commands at the 50 Hz motor-control rate. On Quadruped, \methodname{} increases SR from 50.0\% to 84.3\% and preserves high TCR at 84.7\%. These results indicate that the edit can redirect route-level behavior choices across different mode counts and embodiments, rather than only in the tabletop robot arm manipulation setting.

\begin{table}[hbt]
    \centering
    \footnotesize
    \setlength{\tabcolsep}{1.5pt}
    \renewcommand{\arraystretch}{1.08}
    \setlength{\arrayrulewidth}{0.25pt}
    \caption{
    \textbf{\methodname{} improves deployment success on real-robot tasks while preserving task competence.}
    Mode Change denotes $K \rightarrow |S|$.
    Within each SR or TCR column, entries are Avg / Max over target modes, each target/subset is evaluated with 10 rollouts for the place-bottle task and 20 rollouts for place-knife, block-obstacle, and knife-handover tasks.
    For Place-Bottle $3 \rightarrow 2$, the two evaluated target subsets are $\{\text{left}, \text{right}\}$ and $\{\text{middle}, \text{right}\}$.
    }
    \resizebox{\linewidth}{!}{
        \begin{tabular}{
            @{}l
            !{\vrule width 0.25pt}cc
            !{\vrule width 0.25pt}cc
            !{\vrule width 0.25pt}cc
            !{\vrule width 0.25pt}cc
            !{\vrule width 0.25pt}cc
            @{}
        }
        \toprule
        Task
        & \multicolumn{2}{c}{Place-Knife}
        & \multicolumn{4}{c}{Place-Bottle}
        & \multicolumn{2}{c}{Block-Obstacle}
        & \multicolumn{2}{c}{Knife-Handover(VLA)} \\
        \cmidrule(lr){2-3}
        \cmidrule(lr){4-7}
        \cmidrule(lr){8-9}
        \cmidrule(lr){10-11}
        Mode Change
        & \multicolumn{2}{c}{$2\rightarrow1$}
        & \multicolumn{2}{c}{$3\rightarrow1$}
        & \multicolumn{2}{c}{$3\rightarrow2$}
        & \multicolumn{2}{c}{$2\rightarrow1$}
        & \multicolumn{2}{c}{$2\rightarrow1$} \\
        \cmidrule(lr){2-3}
        \cmidrule(lr){4-5}
        \cmidrule(lr){6-7}
        \cmidrule(lr){8-9}
        \cmidrule(lr){10-11}
        Metric(Avg/Max)
        & SR$\uparrow$ (\%) & TCR$\uparrow$ (\%)
        & SR$\uparrow$ (\%) & TCR$\uparrow$ (\%)
        & SR$\uparrow$ (\%) & TCR$\uparrow$ (\%)
        & SR$\uparrow$ (\%) & TCR$\uparrow$ (\%)
        & SR$\uparrow$ (\%) & TCR$\uparrow$ (\%) \\
        \midrule
        Original policy
        & 40.0 / 70.0 & 80.0 / 80.0
        & 31.7 / 50.0 & 95.0 / 95.0
        & 50.0 / 55.0 & 95.0 / 95.0
        & 47.5 / 50.0 & \textbf{95.0} / 95.0
        & 45.0 / 50.0 & \textbf{90.0} / 90.0 \\
        \rowcolor{oursrow}
        \textbf{\methodname{}}
        & \textbf{80.0 / 85.0} & \textbf{80.0 / 85.0}
        & \textbf{96.7 / 100.0} & \textbf{96.7 / 100.0}
        & \textbf{100.0 / 100.0} & \textbf{100.0 / 100.0}
        & \textbf{92.5 / 100.0} & 92.5 / \textbf{100.0}
        & \textbf{70.0 / 80.0} & 85.0 / \textbf{95.0} \\
        \bottomrule
    \end{tabular}
    }
    \label{tab:real-robot_dp}
\end{table}

\textbf{Transfer to real-robot deployments. (Q4.iv)} Finally, we test \methodname{} on three real-robot tasks using Diffusion Policy and one task using VLA: knife placement, bottle placement, block-obstacle, and knife handover. As shown in Table~\ref{tab:real-robot_dp}, the edited policy shifts rollouts toward the requested behavior mode, increasing SR while preserving task competence. These results show that the same classifier-guided editing objective transfers to physical robot deployments.


\section{Conclusion \& Limitations}
\label{sec:conclusion}
We introduced behavior uncloning, a post-hoc policy-editing problem for
suppressing deployment-undesired modes in already-trained mixed-mode robot
policies. \methodname{} uses a temporary behavior-mode classifier to redirect
undesired-mode rollouts, while a retain loss preserves task competence. The
edited checkpoint discards the classifier and runs through the original
inference path. Across Diffusion Policy and $\pi_{0.5}$, binary and multi-way mode
control, simulation and real robots, \methodname{} improves deployment success
and approaches filtered-data retraining without full retraining or
inference-time steering.

MoRE should be viewed as a mode-control editing method rather than a certified safety layer, and its effectiveness depends on several assumptions. It requires mode-labeled editing data, or closed-loop rollouts whose modes can be reliably inferred, and a classifier interface whose features expose mode information before trajectories have fully committed to a source mode. Extending behavior uncloning to noisier labels, automatically discovered or hierarchical mode taxonomies, longer-horizon tasks, and stronger distribution shifts remains an important direction for future work.

\section{Acknowledgments}
This work was supported in part by GPU hardware and cloud credits provided to the Texas A\&M University PHAI Lab through the NVIDIA Academic Grant Program. We also acknowledge computational resources provided by the Texas A\&M VISION GPU Cluster.

\bibliographystyle{assets/plainnat}
\bibliography{paper}

\clearpage

\appendix

\section{Method and Implementation Details}
\label{app:implementation_details}

\subsection{Classifier Details.}
\label{app:classifier_protocol}

\paragraph{Classifier inputs.}
For VLA policies, we use the representation input
$r_\theta(x)=h_{\theta,t}$, where $h_{\theta,t}$ is the mean-pooled prefix hidden
state of the PaliGemma backbone in $\pi_{0.5}$ at time $t$. Concretely, this is a
$2048$-dimensional vector obtained by averaging all non-padding image-patch and
language-prompt tokens from the last-layer residual stream, after the final RMSNorm
and before the action expert. Classifier training uses cached features
$h_{\theta_0,t}$ from the original mixed policy; editing recomputes
$h_{\theta,t}$ under the current policy so that the redirect gradient can flow
through the editable VLA parameters.

For Diffusion Policy, we use an action-reconstruction input
$r_\theta(x)=(c_t,\hat a_{0,\theta})$. Here $c_t$ denotes the policy condition,
which may include state and image depending on the task. To obtain $\hat a_{0,\theta}$, we sample a diffusion step,
noise the action chunk from $x$, run the denoising UNet, and convert the predicted
noise back to the predicted clean action chunk. Classifier training uses cached
$\hat a_{0,\theta_0}$ from the original policy, editing uses recomputed
$\hat a_{0,\theta}$ with its autograd path into the denoising UNet.

\paragraph{Classifier architecture and training.}
The classifier is a small MLP with three hidden blocks of
width $256$, each consisting of a linear layer, LayerNorm, and a SiLU activation,
followed by a final linear layer that outputs $K$ mode logits. For VLA policies,
the MLP input is the $2048$-dimensional pooled prefix hidden state described
above. For Diffusion Policy, the MLP input is the concatenation of the policy
condition $c_t$ and the flattened clean-action estimate $\hat a_{0,\theta}$.

Classifier parameters are trained only on cached features from the original
mixed policy $\pi_{\theta_0}$ and are then frozen during policy editing. We use
AdamW with batch size $256$ and learning rate $3\times 10^{-4}$ in the default
runs, optimizing a supervised mode-classification objective and selecting the
classifier by held-out validation loss or accuracy.
For Diffusion Policy, including $\hat a_{0,\theta}$ in the classifier input gives
the redirect loss a differentiable path back into the denoising UNet during
policy editing.

\subsection{Classifier-Guidance Derivatives}
\label{app:classifier_guidance_derivatives}

Let $z=g_\phi(r_\theta(x))$ denote the frozen-classifier logits and
$p_j=\operatorname{softmax}(z)_j$. The redirect term in
Eq.~\ref{eq:method_loss} uses the unified subset-probability loss in
Eq.~\ref{eq:subset_redirection_loss} for both single-target and multi-target
editing. Let $p_S=\sum_{i\in S}p_i$. The gradient with respect to
logits is
\begin{equation}
    \frac{\partial \mathcal{L}_{\mathrm{red}}}{\partial z_j}
    =
    p_j
    -
    \mathbf{1}[j\in S]\frac{p_j}{p_S}.
    \label{eq:subset_redirection_grad}
\end{equation}
For a single target $S=\{m^\star\}$, Eq.~\ref{eq:subset_redirection_grad} reduces to
\begin{equation}
    \frac{\partial \mathcal{L}_{\mathrm{red}}}{\partial z_{m^\star}}
    =
    p_{m^\star}-1,
    \qquad
    \frac{\partial \mathcal{L}_{\mathrm{red}}}{\partial z_j}
    =
    p_j
    \quad (j\ne m^\star).
    \label{eq:score_diff_grad}
\end{equation}
Thus gradient descent increases the total probability assigned to the desired
deployment set $S$ and decreases probability mass outside $S$. Since the
classifier is frozen and $z=g_\phi(r_\theta(x))$, the policy update follows the
chain rule
\begin{equation}
    \nabla_\theta \mathcal{L}_{\mathrm{red}}
    =
    \frac{\partial \mathcal{L}_{\mathrm{red}}}{\partial z}
    \frac{\partial g_\phi}{\partial r}
    \frac{\partial r_\theta}{\partial \theta}.
    \label{eq:redirect_chain}
\end{equation}

\section{Additional Per-Target Results}
\label{app:per_target_results}

Tables~\ref{tab:apptemplate_sim_dp}, \ref{tab:apptemplate_sim_dp_baselines},
\ref{tab:apptemplate_sim_vla}, and \ref{tab:apptemplate_real_robot} provide additional per-target breakdowns for the reported MoRE results and, for the simulated Diffusion Policy comparison, the CPL and DynaGuide baselines. The captions specify the rollout counts, target-set definitions, and
which main-paper summaries each table supports.

\subsection{Simulated Diffusion Policy Results}
\label{app:sim_dp_results}

Table~\ref{tab:apptemplate_sim_dp} reports the per-target \methodname{}
breakdown for the simulated Diffusion Policy settings. Table~\ref{tab:apptemplate_sim_dp_baselines}
reports the corresponding per-target CPL~\citep{hejna2024contrastive} and
DynaGuide~\citep{du2026dynaguide} results for the single-target original-demonstration
setting.

\begin{table}[t]
    \centering
    \footnotesize
    \setlength{\tabcolsep}{4pt}
    \setlength{\arrayrulewidth}{0.25pt}
    \renewcommand{\arraystretch}{1.10}
    \begingroup
    \renewcommand{\multirowsetup}{\centering}
    \caption{
    \textbf{Per-target \methodname{} Diffusion Policy results for the simulated benchmark in Tables~\ref{tab:task_suite} and~\ref{tab:pusht_assumptions}.}
    Each row is one target mode under a fixed $K\rightarrow 1$ edit.
    Mode change denotes $K\rightarrow |S|$.
    Push-T (Orig.\ Demos) uses selected original mixed-mode demonstrations as editing data, matching the left block of Table~\ref{tab:pusht_assumptions}; Push-T (Rollout) uses labeled closed-loop rollouts instead.
    Push-Pillars routes are abbreviated as L, LG, RG, and R.
    For \methodname{}, each entry averages over 3 training seeds and 50 initial states per seed (150 rollouts total).
    SR counts only completed rollouts in the desired set $S$; TCR counts task completion regardless of behavior mode.
    }
    \label{tab:apptemplate_sim_dp}
    \begin{tabular*}{\linewidth}{@{\extracolsep{\fill}}>{\centering\arraybackslash}p{1.05in}>{\centering\arraybackslash}p{0.65in}>{\centering\arraybackslash}p{0.85in}*{2}{c}@{}}
        \toprule
        \multirow{2}{*}{Task}
        & \multirow{2}{*}{\shortstack{Mode\\change}}
        & \multirow{2}{*}{Target / $S$}
        & \multicolumn{2}{c}{\methodname{}} \\
        \cmidrule(lr){4-5}
        & & & SR$\uparrow$ (\%) & TCR$\uparrow$ (\%) \\
        \midrule
        \multirow{2}{*}{\shortstack[c]{\scriptsize Push-T\\(Orig.\ Demos)}}
        & $2\rightarrow1$ & left  & 100.0 & 100.0 \\
        & $2\rightarrow1$ & right  & 96.7 & 97.3 \\
        \midrule
        \multirow{2}{*}{\shortstack[c]{\scriptsize Push-T\\(Rollout)}}
        & $2\rightarrow1$ & left   & 78.0 & 78.0 \\
        & $2\rightarrow1$ & right & 98.0 & 98.0  \\
        \midrule
        \multirow{2}{*}{Push-Wall}
        & $2\rightarrow1$ & left & 80.7 & 80.7  \\
        & $2\rightarrow1$ & right & 80.7 & 80.7 \\
        \midrule
        \multirow{2}{*}{Quadruped}
        & $2\rightarrow1$ & left  & 95.3 & 95.3\\
        & $2\rightarrow1$ & right  & 73.3 & 74.0  \\
        \midrule
        \multirow{4}{*}{Push-Pillars}
        & $4\rightarrow1$ & L  & 50.7 & 51.3  \\
        & $4\rightarrow1$ & LG & 96.0 & 99.3  \\
        & $4\rightarrow1$ & RG & 70.7 & 98.7  \\
        & $4\rightarrow1$ & R & 72.0 & 73.3 \\
        \bottomrule
    \end{tabular*}
    \endgroup
\end{table}

\begin{table}[tbh]
    \centering
    \footnotesize
    \setlength{\tabcolsep}{5pt}
    \setlength{\arrayrulewidth}{0.25pt}
    \renewcommand{\arraystretch}{1.10}
    \caption{
    \textbf{Per-target CPL and DynaGuide results underlying the simulated Diffusion Policy comparison in Table~\ref{tab:task_suite}.}
    All rows use single-target $K\rightarrow 1$ editing with original demonstration data.
    Push-T (Orig.\ Demos) rows correspond to the same targets in Table~\ref{tab:apptemplate_sim_dp}; \methodname{} results for those rows appear in that table.
    CPL and DynaGuide were not evaluated on Push-T (Rollout).
    Each entry averages over 3 training seeds and 50 initial states per seed (150 rollouts total).
    }
    \label{tab:apptemplate_sim_dp_baselines}
    \begingroup
    \renewcommand{\multirowsetup}{\centering}
    \begin{tabular}{@{}>{\centering\arraybackslash}p{0.85in}>{\centering\arraybackslash}c>{\centering\arraybackslash}p{0.75in}*{4}{c}@{}}
        \toprule
        \multirow{2}{*}{Task}
        & \multirow{2}{*}{\shortstack{Mode\\change}}
        & \multirow{2}{*}{Target / $S$}
        & \multicolumn{2}{c}{CPL~\citep{hejna2024contrastive}}
        & \multicolumn{2}{c}{DynaGuide~\citep{du2026dynaguide}} \\
        \cmidrule(lr){4-5}\cmidrule(lr){6-7}
        & & & SR$\uparrow$ (\%) & TCR$\uparrow$ (\%) & SR$\uparrow$ (\%) & TCR$\uparrow$ (\%) \\
        \midrule
        \multirow{2}{*}{\shortstack[c]{\scriptsize Push-T\\(Orig.\ Demos)}}
        & $2\rightarrow1$ & left  & 84.7 & 88.0 & 52.7 & 100.0 \\
        & $2\rightarrow1$ & right & 96.0 & 100.0 & 46.7 & 100.0 \\
        \midrule
        \multirow{2}{*}{Push-Wall}
        & $2\rightarrow1$ & left  & 42.7 & 44.7 & 66.7 & 69.3 \\
        & $2\rightarrow1$ & right & 58.7 & 61.3 & 73.3 & 73.3 \\
        \midrule
        \multirow{4}{*}{Push-Pillars}
        & $4\rightarrow1$ & L & 32.0 & 34.7 & 37.3 & 66.0 \\
        & $4\rightarrow1$ & LG & 62.7 & 66.0 & 40.7 & 58.0 \\
        & $4\rightarrow1$ & RG & 47.3 & 70.7 & 14.7 & 60.0 \\
        & $4\rightarrow1$ & R & 56.0 & 56.0 & 45.3 & 62.0 \\
        \midrule
        \multirow{2}{*}{Quadruped}
        & $2\rightarrow1$ & left  & 77.3 & 98.0 & 54.7 & 100.0 \\
        & $2\rightarrow1$ & right & 58.0 & 63.3 & 46.0 & 100.0 \\
        \bottomrule
    \end{tabular}
    \endgroup
\end{table}

\subsection{Simulated VLA Results}
\label{app:sim_vla_results}

Table~\ref{tab:apptemplate_sim_vla} reports the per-target and multi-target
\methodname{} breakdowns for the $\pi_{0.5}$ VLA settings in
Table~\ref{tab:vla_generalization}.

\begin{table}[tbh]
    \centering
    \footnotesize
    \setlength{\tabcolsep}{5pt}
    \setlength{\arrayrulewidth}{0.25pt}
    \renewcommand{\arraystretch}{1.10}
    \begingroup
    \renewcommand{\multirowsetup}{\centering}
    \caption{
    \textbf{Per-target \methodname{} VLA results underlying the Avg/Max summaries in Table~\ref{tab:vla_generalization}.}
    Mode change denotes $K\rightarrow |S|$.
    Push-Pillars routes are abbreviated as L, LG, RG, and R; multi-target rows list the desired mode set $S$.
    For multi-target edits, SR treats any completed rollout whose mode lies in $S$ as mode-aligned.
    For Push-Pillars $4\rightarrow 2$ and $4\rightarrow 3$, main-table Avg/Max averages over the evaluated target subsets listed here.
    Each setting uses 50 evaluation seeds.
    }
    \label{tab:apptemplate_sim_vla}
    \begin{adjustbox}{max width=\linewidth}
    \begin{tabular}{@{}>{\centering\arraybackslash}p{0.95in}>{\centering\arraybackslash}p{0.75in}>{\centering\arraybackslash}p{1.05in}*{2}{c}@{}}
        \toprule
        \multirow{2}{*}{Task}
        & \multirow{2}{*}{Mode change}
        & \multirow{2}{*}{Target / $S$}
        & \multicolumn{2}{c}{\methodname{}} \\
        \cmidrule(lr){4-5}
        & & & SR$\uparrow$ (\%) & TCR$\uparrow$ (\%) \\
        \midrule
        \multirow{2}{*}{Push-Wall}
        & $2\rightarrow1$ & left  & 64.0 & 74.0  \\
        & $2\rightarrow1$ & right & 80.0 & 84.0  \\
        \midrule
        \multirow{14}{*}{Push-Pillars}
        & $4\rightarrow1$ & L & 74.0 & 80.0  \\
        & $4\rightarrow1$ & LG & 66.0 & 100.0  \\
        & $4\rightarrow1$ & RG & 42.0 & 88.0 \\
        & $4\rightarrow1$ & R & 54.0 & 80.0 \\
        & $4\rightarrow2$ & \{L, LG\} & 70.0 & 90.0  \\
        & $4\rightarrow2$ & \{L, RG\} & 60.0 & 94.0  \\
        & $4\rightarrow2$ & \{L, R\} & 72.0 & 84.0 \\
        & $4\rightarrow2$ & \{LG, RG\} & 76.0 & 88.0 \\
        & $4\rightarrow2$ & \{LG, R\} & 70.0 & 90.0  \\
        & $4\rightarrow2$ & \{RG, R\} & 74.0 & 82.0  \\
        & $4\rightarrow3$ & \{L, LG, RG\} & 76.0 & 86.0 \\
        & $4\rightarrow3$ & \{L, LG, R\} & 76.0 & 86.0 \\
        & $4\rightarrow3$ & \{L, RG, R\} & 66.0 & 82.0  \\
        & $4\rightarrow3$ & \{LG, RG, R\} & 70.0 & 76.0 \\
        \bottomrule
    \end{tabular}
    \end{adjustbox}
    \endgroup
\end{table}

\subsection{Real-Robot Deployment Results}
\label{app:real_robot_results}

Table~\ref{tab:apptemplate_real_robot} reports the real-robot per-target
breakdowns underlying Table~\ref{tab:real-robot_dp}.

\begin{table}[tbh]
    \centering
    \footnotesize
    \setlength{\tabcolsep}{6pt}
    \setlength{\arrayrulewidth}{0.25pt}
    \renewcommand{\arraystretch}{1.10}
    \begingroup
    \renewcommand{\multirowsetup}{\centering}
    \caption{
    \textbf{Per-target \methodname{} real-robot results underlying the Avg/Max summaries in Table~\ref{tab:real-robot_dp}.}
    Mode change denotes $K\rightarrow |S|$.
    Place-Knife, Place-Bottle, and Block-Obstacle use Diffusion Policy; Knife-Handover uses the $\pi_{0.5}$ VLA backbone.
    Place-Bottle uses 10 rollouts per target mode or subset; the other tasks use 20 rollouts per target mode or subset.
    For Place-Bottle $3\rightarrow 2$, the two evaluated target subsets are $\{\text{left}, \text{right}\}$ and $\{\text{middle}, \text{right}\}$, matching Table~\ref{tab:real-robot_dp}.
    On Knife-Handover, each row fixes a different desired set $S$ (handle-first or blade-first).
    }
    \label{tab:apptemplate_real_robot}
    \begin{adjustbox}{max width=\linewidth}
    \begin{tabular}{@{}>{\centering\arraybackslash}p{1.0in}>{\centering\arraybackslash}c>{\centering\arraybackslash}p{1.05in}*{2}{c}@{}}
        \toprule
        \multirow{2}{*}{Task}
        & \multirow{2}{*}{Mode change}
        & \multirow{2}{*}{Target / $S$}
        & \multicolumn{2}{c}{\methodname{}} \\
        \cmidrule(lr){4-5}
        & & & SR$\uparrow$ (\%) & TCR$\uparrow$ (\%) \\
        \midrule
        \multirow{2}{*}{Place-Knife}
        & $2\rightarrow1$ & left (hand)  & 85.0 & 85.0 \\
        & $2\rightarrow1$ & right (box) & 75.0 & 75.0 \\
        \midrule
        \multirow{5}{*}{Place-Bottle}
        & $3\rightarrow1$ & left & 90.0 & 90.0 \\
        & $3\rightarrow1$ & middle & 100.0 & 100.0 \\
        & $3\rightarrow1$ & right & 100.0 & 100.0 \\
        & $3\rightarrow2$ & \{left, right\} & 100.0 & 100.0 \\
        & $3\rightarrow2$ & \{middle, right\} & 100.0 & 100.0 \\
        \midrule
        \multirow{2}{*}{Block-Obstacle}
        & $2\rightarrow1$ & left  & 85.0 & 85.0 \\
        & $2\rightarrow1$ & right & 100.0 & 100.0 \\
        \midrule
        \multirow{2}{*}{Knife-Handover}
        & $2\rightarrow1$ & handle-first & 60.0 & 75.0 \\
        & $2\rightarrow1$ & blade-first  & 80.0 & 95.0 \\
        \bottomrule
    \end{tabular}
    \end{adjustbox}
    \endgroup
\end{table}

\end{document}

%% file: math_commands.tex
\usepackage{amsmath,amsfonts,bm}

\def\eqref#1{equation~\ref{#1}}

\def\1{\bm{1}}

\DeclareMathAlphabet{\mathsfit}{\encodingdefault}{\sfdefault}{m}{sl}
\SetMathAlphabet{\mathsfit}{bold}{\encodingdefault}{\sfdefault}{bx}{n}

